\documentclass[sigconf]{acmart}

\copyrightyear{2026}
\acmYear{2026}
\setcopyright{cc}
\setcctype{by}
\acmConference[WWW Companion '26] {Companion Proceedings of the ACM Web Conference 2026}{April 13--17, 2026}{Dubai, United Arab Emirates.}
\acmBooktitle{Companion Proceedings of the ACM Web Conference 2026 (WWW Companion '26), April 13--17, 2026, Dubai, United Arab Emirates}
\acmISBN{979-8-4007-2308-7/2026/04}
\acmDOI{10.1145/3774905.3795092}

\usepackage{graphicx}
\usepackage{listings}
\usepackage{multirow}
\usepackage{multicol}

\lstdefinelanguage{json}{}
\AtBeginDocument{%
  }

\begin{document}

\title[A Prompt-Aware Structuring Framework for Reliable Reuse of AI-Generated Content \\ in the Agentic Web]{A Prompt-Aware Structuring Framework for Reliable Reuse of AI-Generated Content in the Agentic Web}


\author{Shusaku Egami}
\orcid{0000-0002-3821-6507}
\email{s-egami@aist.go.jp}
\affiliation{%
  \institution{National Institute of Advanced Industrial Sciencen and Technology (AIST)}
  \city{Koto}
  \state{Tokyo}
  \country{Japan}
}
\author{Masahiro Hamasaki}
\orcid{0000-0003-3085-7446}
\email{masahiro.hamasaki@aist.go.jp}
\affiliation{%
  \institution{National Institute of Advanced Industrial Sciencen and Technology (AIST)}
  \city{Koto}
  \state{Tokyo}
  \country{Japan}
}

\renewcommand{\shortauthors}{Shusaku Egami and Masahiro Hamasaki}

\begin{abstract}

The evolution of Large Language Models (LLMs) and the software agents built on them (AI agents) marks a turning point in the transition from a human-centric Web to an ``Agentic Web'' driven by AI agents. However, for AI-Generated Content (AIGC), which is expected to dominate the Web, there is currently no mechanism for agents to verify its reliability, reproducibility, or license compliance during generation. This lack of transparency risks causing chained hallucinations and compliance violations through the reuse of AIGC. Consequently, a framework to manage the provenance and generation conditions of AIGC is essential. 
In this paper, we present a framework that automatically attaches structured metadata to AIGC at generation time, including modularized prompts, contexts, thoughts, model information, hyperparameters, and confidence. The metadata is enveloped together with verifiable credentials to support the reliable assessment and reuse of AIGC. This framework enables efficient curation of structured AIGC and facilitates its safe use for applications such as fine-tuning and knowledge distillation.

\end{abstract}




\maketitle

\section{Introduction}

With the rapid advancement of large language models (LLMs) and AI agents built upon them, the Web is at a turning point: it is shifting from a space where humans primarily read and write content to an ``Agentic Web,'' in which AI agents autonomously perform these tasks on behalf of users.
AI-generated content (AIGC) is already widespread across the Web, and a recent analysis revealed that 74\% of newly detected webpages by web crawlers in April 2025 contained AIGC~\cite{soulo_74_2025}.

However, there is currently no mechanism that allows AI agents to reliably assess the trustworthiness or reproducibility of AIGC—issues that are inherent to such content. When hallucinations are present in AIGC, these errors can be propagated and amplified across the Web through AI agents. Moreover, the use of low-quality AIGC for fine-tuning and knowledge distillation can degrade the performance of AI agents, thereby creating a negative feedback loop within the Agentic Web ecosystem.

Therefore, it is essential to explicitly manage the provenance and generation conditions of AIGC, and to establish mechanisms that enable verification of content quality and reliability when AIGC is reused.

\begin{figure*}[h]
  \centering
  \includegraphics[width=\linewidth]{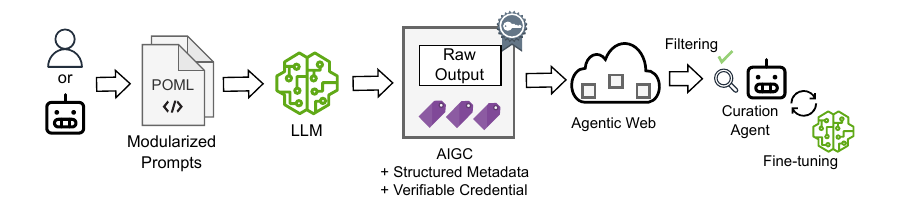}
  \vspace{-10mm}
  \caption{Overview of the proposed framework}
  \label{fig:overview}
  \vspace{-2mm}
\end{figure*}

In this study, we propose a framework for publishing AI-generated content (AIGC) as verifiable and reusable knowledge resources by automatically attaching structured metadata at generation time. The metadata explicitly represents not only model outputs but also modularized prompts, generation contexts, thoughts, model information, and hyperparameters, and is enveloped together with verifiable credentials to ensure credibility and provenance. Based on this structured representation, we curate AIGC by mechanically evaluating instruction-following fidelity using prompt-level requirements, and examine whether the curated AIGC can be safely reused for model fine-tuning. Specifically, we conduct instruction-following fine-tuning on ComplexBench~\cite{wen_benchmarking_2024} by training a small student model using AIGC generated by multiple teacher models. Experimental results demonstrate that metadata-driven AIGC curation consistently improves the requirements following ratio compared to the baseline.
Figure~\ref{fig:overview} provides an overview of the proposed framework and evaluation.
These findings suggest that explicitly managing the prompt structures and generation processes of AIGC is essential for its safe and effective reuse in the era of the agentic web.

\section{Related Work}


A new Internet paradigm, the Agentic Web~\cite{yang_agentic_2025}, has been introduced in which AI agents powered by LLMs autonomously perform tasks. In the era of the Agentic Web, the Web is no longer a human-centered space for browsing; instead, it evolves into an agent-driven environment where intelligent agents execute complex tasks on behalf of users. 
As enabling technologies for such a Web, agent protocols such as Model Context Protocol (MCP)\footnote{\url{https://modelcontextprotocol.io/}} and Agent-to-Agent (A2A)\footnote{\url{https://a2a-protocol.org/latest/}} play a crucial role in facilitating coordination and collaboration among diverse services and agents. MCP aims to provide a standardized protocol for connecting LLM applications to external systems. Numerous MCP servers and their lists are already available~\cite{hou_model_2025}.
A2A is designed to offer a standard mechanism for cooperation among heterogeneous AI agents. However, despite these advances, mechanisms for safely reusing content generated by AI agents in the era of the Agentic Web remain insufficiently developed.

Prompts serve as the primary programming interface in LLM applications, directly shaping LLM behavior and outputs. In agent-to-agent interactions, prompts further become the primary communication unit.
POML~\cite{zhang_prompt_2025} is a markup language for prompt orchestration that supports structured prompt markup and template engines. This enables improved reusability and maintainability of prompts, as well as data-driven and dynamic prompt generation.
Meanwhile, in agent-to-agent interactions, the content generated by an LLM can also become part of a prompt and therefore constitutes a primary communication unit. Therefore, in the Agentic Web era, not only the reusability of prompts created by humans but also the reusability of outputs generated by prompts is considered an important element.

As AIGC proliferates on the Web, it becomes increasingly important to scrutinize its credibility. The Verifiable Credentials Data Model (VCDM)\footnote{\url{https://www.w3.org/TR/vc-data-model-2.0/}}, a W3C Recommendation, provides a means to verify information and establishes the foundation of trust. Verifiable Credentials (VCs) are tamper-resistant credentials that creators can verify using cryptography. Furthermore, within VCDM, Decentralized Identifiers (DIDs)\footnote{\url{https://www.w3.org/TR/did-1.0/}}, also a W3C Recommendation, are used to identify issuers and presenters. Utilizing VCDM enables the distribution of verifiable digital certificates confirming both the content creator's identity and the content's integrity. This makes VCDM a crucial technology that supports the Agentic Web.

\section{Publishing AI-generated Content with Structured Metadata}

\subsection{Metadata Schema}
\label{sec:schema}

\begin{figure}[h]
  \centering
  \includegraphics[width=\linewidth]{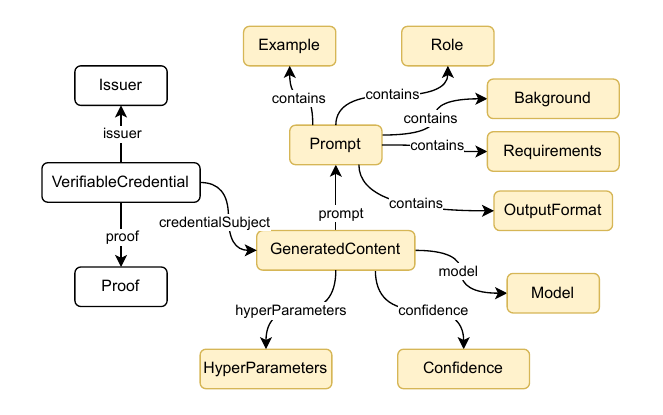}
  \vspace{-10mm}
  \caption{Proposed metadata schema}
  \label{fig:schema}
  \vspace{-10mm}
\end{figure}

We created the competency questions (CQs) to design the metadata schema.

\begin{itemize}
    \item [CQ1] What prompts (instructions, role, background information, output format, etc.) were used to generate this AIGC?
    \item [CQ2] Who created the prompts to generate this AIGC?
    \item [CQ3] Which model/agent was used to generate this AIGC?
    \item [CQ4] Which hyperparameters were used to generate this AIGC?
    \item [CQ5] With what confidence was this AIGC generated?
    \item [CQ6] When was this AIGC generated?
    \item [CQ7] Who modified this AIGC?
    \item [CQ8] What was the reasoning process used to generate this AIGC?
    \item [CQ9] What is the digital signature of this AIGC?
\end{itemize}

We listed the necessary classes and designed the metadata schema based on these CQs.
Figure~\ref{fig:schema} shows the schema diagram of the metadata to be attached to AIGC.
A distinctive feature is that the prompts used for generation are included in the metadata in a structured format. This makes it easier to verify whether the AIGC appropriately follows the prompt instructions.
JSON-LD is used as the metadata format.

\subsection{Modularizing Prompt}

Prompts have typically been created on the fly or maintained as plain text, but the use of prompt structuring techniques is expected to become more widespread in the future. 
In this study, we assume an environment in which prompts are modularized, and we construct a POML dataset in which each prompt is decomposed into the modules Role, Background, Requirements, Example, and OutputFormat.

We use ComplexBench~\cite{wen_benchmarking_2024} as the source dataset. ComplexBench is a benchmark designed to evaluate the instruction-following ability of LLMs, and its prompts contain a wide variety of instructions. Using an LLM, we extract from each prompt the elements corresponding to Role, Background, Requirements, Example, and OutputFormat, and generate separate POML files for each module. The main POML file is configured to import these module files so that rendering the POML produces a complete prompt in Markdown format.

We use GPT-5-mini as the LLM for this extraction process.
In total, we created 1,150 main POML files and 5,024 POML files including all modular components.

\subsection{Generating AIGC with Structured Metadata}

Based on the schema defined in Section~\ref{sec:schema}, we encapsulate the model outputs, prompts, and all associated metadata in JSON-LD format.
By modularizing prompts using POML, the structured prompts can be directly incorporated as metadata for the AIGC. When prompts are not managed in a modularized form, they can also be structured using an LLM following the same procedure described in the previous section.

The Confidence field contains the logits produced during model inference. We obtain the log probabilities for each output token and then compute an overall confidence score for the entire output.
When inference models generate explicit chain-of-thought outputs, the generated content may include internal reasoning traces (e.g., ``<think></think>'').
We extract this thought using a lightweight, tag-based rule-matching method and attach it as optional metadata when the tags are present.

To ensure that the published AIGCs have not been tampered with and that the creator (issuer) information is authentic, we issue a verifiable credential. The generated content and all metadata are normalized and hashed, and then signed using the creator’s private key.
The signature value, signature method, signature date, and issuer information are described using the vocabulary defined in the Verifiable Credential data model.

An example of the generated structured AIGC JSON-LD is shown as follows.

\begin{lstlisting}[]
{ "@context": [
    "https://www.w3.org/ns/credentials/v2",
    "https://w3id.org/security/data-integrity/v1",
    ...
  ],
  "id": "urn:uuid:f5c4c481-7915-441e-9c21-672ad62e12f3",
  "type": [
    "VerifiableCredential",
    "AIGCContentCredential"
  ],
  "issuer": {
    "id": "did:web:ease112.github.io",
    "name": "Shusaku Egami"
  },
  "validFrom": "2025-12-10T01:17:04Z",
  "credentialSubject": {
    "@id": "_:N973b9cd27fcd479296b6962141cb9053",
    "@type": "GeneratedContent",
    "confidence": {
      "@id": "_:Ne21d5148d8334ea2b5365cd8feb30a2b"
    },
    "hyperParameter": {
      "@id": "_:Nd7130c293157420ea963659a49a82a9b"
    },
    "label": "analysis...",
    "model": {
      "@id": "https://huggingface.co/openai/gpt-oss-20b"
    },
    "prompt": {
      "@id": "_:N67dea220da2c47efae8fa318b66bf904"
    },
    "value": "analysisWe need to write a news article summarizing the financing round. ..."
  },
  "@graph": [
    {
      "@id": "_:N67dea220da2c47efae8fa318b66bf904",
      "@type": "Prompt",
      "contains": [
        {
          "@id": "_:N831532dea6454cf38ef0e8c23e32b861"
        },
        ...
      ],
      "value": "# Role\n\nYou are an assistant for Practical Writing tasks.\n\n# ..."
    },
    {
      "@id": "_:N831532dea6454cf38ef0e8c23e32b861",
      "@type": "Role",
      "value": "You are an assistant for Practical Writing tasks."
    },
    ...
    {
      "@id": "_:Ne21d5148d8334ea2b5365cd8feb30a2b",
      "mean": -0.4548458994601176,
      ...
    },
    {
      "@id": "_:Nd7130c293157420ea963659a49a82a9b",
      "max_tokens": 2000,
      "temperature": 1.0
    },
    {
      "@id": "https://huggingface.co/openai/gpt-oss-20b",
      "@type": "Model",
      "label": "openai/gpt-oss-20b"
    }
  ],
  "proof": {
    "type": "DataIntegrityProof",
    "created": "2025-12-10T01:17:04Z",
    "proofPurpose": "assertionMethod",
    "verificationMethod": "did:web:ease112.github.io#key-1",
    "cryptosuite": "eddsa-rdfc-2022",
    "proofValue": "U7wKOddv..."
  }}
\end{lstlisting}

\subsection{Curation}

We consider a scenario in which AIGC is reused for model fine-tuning and knowledge distillation.
A curation agent retrieves the metadata associated with each AIGC instance and evaluates its quality. Although various quality metrics can be considered, we focus on instruction-following fidelity in this study.

The curation agent accesses the values of the modularized prompt components included in the metadata and checks whether the generated content satisfies the instructions specified in each prompt module.
If the AIGC is judged to satisfy all instructions, the AIGC and the corresponding prompt are reused for fine-tuning.

\section{Experiments}

This experiment demonstrates that the structured metadata generated by the proposed framework enables efficient data curation for high-quality AIGC and is useful for reuse in tasks such as LLM distillation.

\subsection{Experimental Setup}

\subsubsection{Tasks}
The objective of this experiment is to evaluate the instruction-following capability of LLMs, focusing on adherence to constraints rather than generation quality. We use the ComplexBench dataset, split into 68\% for training and 32\% for testing.
\subsubsection{Models}
In this experiment, we finetune a small-scale student model using structured AIGC generated by multiple teacher models with large parameters. Specifically, Qwen3-32B~\cite{yang_qwen3_2025}, gpt-oss-20b~\cite{openai_gpt-oss-120b_2025}, and Llama-3.1-8B-Instruct~\cite{grattafiori_llama_2024} are used as the teacher models, while Llama-3.2-1B-Instruct~\cite{noauthor_llama_nodate} is used as the student model.

\subsubsection{Training strategies}
For the training set, we first generate structured AIGCs from each teacher model using our proposed method, resulting in three distinct outputs per prompt. We compare two fine-tuning strategies:
\begin{itemize}
    \item Random-FT (baseline): The student model is fine-tuned on data randomly selected from the three teacher outputs.
    \item Curated-FT: The student model is fine-tuned on the single best output selected based on a quality evaluation of the three teacher outputs.
\end{itemize}

\subsubsection{Evaluation}
We evaluate the outputs generated by the Random-FT and Curated-FT models using prompts from the test split. To measure instruction-following capability, we use the specific questions associated with each prompt provided by ComplexBench. We employ GPT-5.1-mini as the evaluator model.

We adopt two metrics: Requirements Following Ratio (RFR) and Full Requirements Following Ratio (FRFR). 
RFR is calculated as the ratio of questions answered ``yes'' out of the total number of questions across the entire dataset. In contrast, FRFR is the ratio of prompts for which the evaluator answers ``yes'' to all corresponding questions (i.e., strict prompt-level success rate).

\subsection{Experimental Results}

Table~\ref{table:results} shows the experimental results. For reference, the performance of the teacher models is also included. The results show that Curated-FT outperformed the baseline Random-FT in both RFR and FRFR metrics. These findings demonstrate that the structured AIGC generated by our proposed framework enables effective data curation for model fine-tuning and knowledge distillation.

\begin{table}[t]
\setlength{\tabcolsep}{2pt} 
  \renewcommand{\arraystretch}{0.9}
  \caption{Requirements following ratios on ComplexBench}
  \label{table:results}
  \vspace{-1.0em}
  \centering
  \begin{tabular}{llcc}
    \hline
    \multicolumn{2}{c}{Model} & RFR (\%) & FRFR (\%) \\ \hline
    \multirow{3}{*}{\begin{tabular}[c]{@{}l@{}}Teacher\\ LLM\end{tabular}} & Llama-3.1-8B-Instruct & 63.02 & 33.24 \\
                             & gpt-oss-20b           & 61.17 & 37.67 \\
                             & Qwen3-32B             & 65.02 & 39.89 \\ \hline
    \multirow{3}{*}{\begin{tabular}[c]{@{}l@{}}Student\\ LLM\end{tabular}} & Llama-3.2-1B-Instruct              & 44.91 & 13.57 \\
                             & Llama-3.2-1B-Instruct (Random-FT)  & 45.45 & 15.79 \\
                             & Llama-3.2-1B-Instruct (Curated-FT) & \textbf{47.30} & \textbf{16.07} \\ \hline
  \end{tabular}
  \vspace{-5mm}
\end{table}

\section{Discussion}

Our proposed framework assumes prompt decomposition into modules, such as role, background, requirements, example, and output format, using POML. This design enables curation agents to mechanically evaluate which requirements are satisfied and which are violated, rather than relying on ambiguous judgments about whether the generated result seems good.

The experimental results suggest that explicitly managing prompt structures and generation processes during AIGC reuse has a direct impact on the learning quality of the core models within AI agents.
Therefore, this study indicates that retaining not only the model outputs but also their associated context and provenance is beneficial for AIGC reuse, supporting the potential for AIGC to function as a structured and contextualized knowledge resource.
In interpreting these results, it is important to note that the evaluation relies on an LLM-based judge without calibration or agreement checks, and our experimental evaluation focuses on measuring relative improvements under a fixed evaluation protocol rather than absolute correctness.

On the other hand, several challenges remain in this study. In particular, sufficient motivation has not yet been established for AIGC issuers to actively publish structured and contextualized AIGCs using our framework. While the structured AIGCs generated by our framework contribute to improved reusability and reliability, these benefits remain primarily indirect advantages for issuers. Consequently, for issuers prioritizing short-term, direct gains, this may not constitute a strong incentive for adoption. A key future challenge is to present application examples and usage scenarios that can provide clearer incentives for AIGC issuers.

\section{Conclusion}
This paper proposed a framework for structuring AI-generated content (AIGC) to enable reliable and verifiable reuse in the Agentic Web.
By automatically attaching structured metadata—such as modularized prompts, generation contexts, thoughts, model information, and hyperparameters—and issuing verifiable credentials, the proposed approach enables reliable assessment and safe reuse of AIGC by AI agents. 
Experimental results on instruction-following fine-tuning demonstrated that metadata-based curation consistently outperforms random data selection, indicating that structured AIGC can serve as higher-quality training data for downstream tasks. 
These findings suggest that explicitly managing the provenance and generation conditions of AIGC is a key step toward building a sustainable and trustworthy Agentic Web.
While our current experiments focus on fine-tuning, future work will evaluate the effectiveness of VCs by introducing malicious agents into the environment.

\section*{Acknowledgements}
This paper is based on results obtained from a project, JPNP25006, commissioned by the New Energy and Industrial Technology Development Organization (NEDO), and JSPS KAKENHI Grant Numbers JP23H03688 and JP25K03232.

\bibliographystyle{ACM-Reference-Format}
\bibliography{ref}

\end{document}